\documentclass{article}
\usepackage{nips14submit_e,times}
\usepackage{hyperref}
\usepackage{url}
\usepackage{amsfonts, amsmath, amssymb}
\usepackage{epsfig, subfigure, subfloat, graphicx}

\title{Unsupervised Learning For Effective User Engagement on Social Media}

\author{
Thai T. Pham \\
\texttt{thaipham@stanford.edu} \\
\And
Camelia Simoiu \\
\texttt{csimoiu@stanford.edu} 
}

\nipsfinalcopy

\begin{document}

\maketitle

\begin{abstract}
In this paper, we investigate the effectiveness of unsupervised feature learning techniques in predicting user engagement on social media. Specifically, we compare two methods to predict the number of feedbacks (i.e., comments) that a blog post is likely to receive. We compare Principal Component Analysis (PCA) and sparse Autoencoder to a baseline method where the data are only centered and scaled, on each of two models: Linear Regression and Regression Tree. We find that unsupervised learning techniques significantly improve the prediction accuracy on both models. For the Linear Regression model, sparse Autoencoder achieves the best result, with an improvement in the root mean squared error (RMSE) on the test set of $42 \%$ over the baseline method. For the Regression Tree model, PCA achieves the best result, with an improvement in RMSE of $15 \%$ over the baseline. 

\end{abstract}

\section{Introduction}

Social media has become an important tool for public engagement. Businesses are interested in engaging with potential buyers on platforms such as Facebook or their company news pages; bloggers are publishers are interested in increasing their follower and reader base by writing captivating articles that prompt high levels of user feedback (in the form of likes, comments, etc.). Such groups will inevitably need to understand the types of content that is likely to elicit the most user engagement as well as any underlying patterns in the data such as temporal trends.  Specifically, we focus on predicting the number of feedbacks (i.e., comments) that a blog post is expected to receive. Given a set of blog documents that appeared in the past, for which the number and time stamp received, the task is to predict how many feedbacks recently published blog-entries will receive in the next $H$ hours.

A first challenge in answering this question is how to effectively pre-process the data. Despite offering a rich source of information, such data sets are usually noisy, high-dimensional, with many correlated features. In this paper, we focus on two unsupervised learning approaches for pre-processing: the traditional method of Principal Component Analysis (PCA), and a more recently-developed method from deep-learning, the sparse Autoencoder. These pre-processing methods are generally used to reduce dimensionality, eliminate correlations among variables (since there may be a number of irrelevant features in the data set), decrease the computation time, and extract good features for subsequent analyses. PCA linearly transforms the original inputs into new uncorrelated features. The sparse Autoencoder is trained to reconstruct its own inputs through a non-recurrent neural network, and it creates as output new features from non-linear combinations of the old ones. We compare the effects of these two methods on two prediction models of linear regression and regression trees in the feedback prediction task. 

The rest of the paper proceeds as follows. Section $2$ includes a short review of related literature. Section $3$ describes the data set used. Section $4$ discusses the unsupervised feature learning methods of PCA and sparse Autoencoder. Section $5$ gives comparative results from different models for the unprocessed data, and the pre-processed data using PCA and sparse Autoencoder. Section $6$ concludes and opens future research directions. Section $7$ acknowledges help.   

\section{Literature Review}

Cao et. al. $[2]$ compare PCA, Kernel PCA (KPCA), and Independent Component Analysis (ICA) as applied to Support Vector Machine (SVM) for feature extraction to three data sets (sunspot data, Satan Fe data set A, and five real futures contracts). SVM by feature extraction using PCA, KPCA or ICA can achieve better generalization performance than that without feature extraction, and that KPCA and ICA perform better than PCA on all the studied data sets, with the best performance in KPCA. The reason lies in the fact that KPCA and ICA can explore higher order information of the original inputs than PCA. Based on these results, we expect the (sparse) Autoencoder to perform better than PCA as it is also a non-linear combination of features.

Buza $[1]$ uses the same data set as we do and compares a variety of models to predict the number of future feedbacks for a blog. They consider two performance metrics: Area Under Curve explained (AUC), and the number of blog pages that were predicted to have the largest number of feedbacks out of the top ten blog pages that had the highest number of feedbacks in reality. Buza $[1]$ examines various models: a multilayer perceptron model, RBF-networks, regression trees (REP-tree, M5P-tree), nearest neighbor models, multivariate linear regression and bagging, however do not pre-process the data. He finds that M5P Trees and REP Trees seem to work very well both in terms of accuracy and runtime, averaging $5 - 6$ feedbacks and $84 - 92 \%$ for the examined models. The neural network model is competitive to the regression trees. The best models are the linear model (AUC $= 92 \%$ and $5.2$ hits) and neural network (AUC $= 87 \%$, $5.8$ hits). Bagging does not statistically improve the performance of MLPs and RBF-Network both in terms of comments and AUC (improved the absolute performance, but not taking into account standard deviations). We use the same data set as in Buza $[1]$, but approach the problem differently. Specifically, we use unsupervised feature learning techniques to pre-process the data before running prediction models. 

\section{Data Set}

This data set is made available from the UCI Machine Learning Repository, and comprises a total of $37,279$ crawled blog pages. The prediction task is to predict the number of comments for a blog post in the upcoming $H = 24$ hours. The processed data has a total of $280$ features (without the target variable, i.e., number of feedbacks). \footnote{We use the processed data provided in $[1]$.} The blog documents (instances) are transformed into vectors to be inputted to the machine learning algorithm. This collection corresponds approximately $6$ GB of plain HTML document (i.e., without images). The following features are extracted from each document: 

\begin{itemize}
    \item Basic features: Number of links and feedbacks in the previous $24$ hours; number of links and feedbacks in the previous $48$ hours; how the number of links and feedbacks increased/decreased since to the publication date of the blog; number of links and feedbacks in the first $24$ hours after the publication of the document.
    
    \item Textual features: The most discriminative bag-of-words features.
    
    \item Weekday features: Binary indicator features that describe on which day of the week the main text of the document was published and for which day of the week the prediction has to be calculated.
    
    \item Parent features: A document dP is treated as a patent of document d, if d is a reply to dP, i.e., there is a trackback link on dP that points to d; parent features are the number of parents, minimum, maximum and average number of feedbacks that the parents received.
\end{itemize}

 \section{Unsupervised Feature Learning Methods}

\begin{figure}[ht!]
     \begin{center}
        \subfigure[]{
            \includegraphics[width=0.45\textwidth]{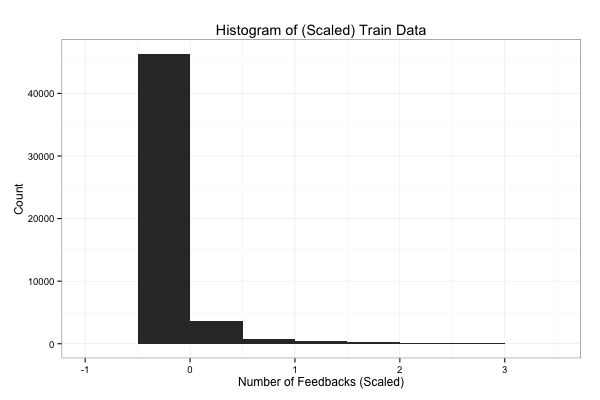}
            \label{hist_train}
        }
        \subfigure[]{
           \includegraphics[width=0.45\textwidth]{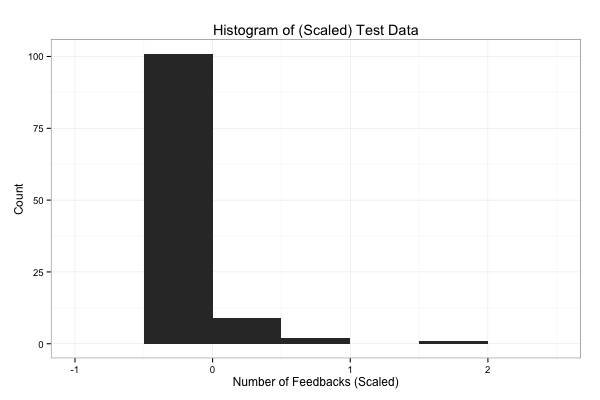}
           \label{hist_test}
		}
    \end{center}
    \caption{Histograms of The Output Variable For Training and Test Sets.}
\end{figure}

Prior to implementing any unsupervised feature learning method, we first eliminate four variables which contain only zeros, resulting in $276$ predictor variables (i.e., features) and one outcome variable. Out of these variables, $58$ are continuous, and the remainder are binary. We center and scale all non-binary variables to have zero mean and unit standard deviation. This step prevents any one variable from dominating the variance of the data set simply due to taking values in a larger range. We also center and scale the outcome variable for testing later. Histograms of the number of feedbacks received for both train and test sets are included in Figures \ref{hist_train} and \ref{hist_test}.

We observe that most values of the centered and scaled response variable are concentrated near zero for both training and test sets. This prevents any one point from potentially skewing the estimation results later on. 

\subsection{Principal Component Analysis}

PCA is a powerful technique for extracting structure from high-dimensional data sets. We implement PCA in order to capture the intrinsic variability in the data and reduce the dimensionality of the data set. An important decision in PCA analysis is to determine the optimal number of principal components to use, since the data will be projected onto this new basis of principal components before any subsequent analysis. To determine the optimal number of components, we implement two Gap-style tests as in $[3]$. In both tests, the Gap formula has the same form

\begin{equation*}
    Gap(k) = E[log(W_k)] - log(W_k),
\end{equation*}

where $log(W_k)$ is the logarithm of some objective function corresponding to the first $k$ principal components and $E[log(W_k)]$ is calculated in the similar way with the expectation taken over uniform samples from the smallest subspace containing the original data.

\subsubsection{Gap Test Using Reconstruction Error}

In the first approach, we define $W_k$ to be the reconstruction error in using the first $k$ principal components to approximate the data. Specifically, let $\{x_{ij}^k\}$, $j = 1, ..., 276$ be the projection of the $i^{th}$ data point onto the rank $k$ principal component approximation. Then $W_k$ is defined by
\begin{equation*} W_k = \sum_{i, j} (x_{ij} - \widehat{x}_{ij}^k)^2 \end{equation*}

The steps for this Gap-style test are as follows.

\begin{itemize}
	
	\item[i.] After calculating the $276$ principal components, for each $k \in \{1, ..., 276\}$, we determine $W_k$ as above and take their log values to obtain $276$ values, $log (W_k)$'s.

	\item[ii.] We generate $B$ samples uniformly from the smallest subspace containing the original data and find the principal components for this newly generated matrix. As above, for each $k \in \{1, ..., 276\}$, we calculate $W_k$ and take their log values. So for each $k$, we obtain $B$ values; we take their average to get an estimate for $E[log (W_k)]$ and calculate their standard deviation. We obtain $276$ such values, and call them $sd[k]$ for $k \in \{1, ..., 276\}$. 

	\item[iii.] Calculate the adjusted standard error according to the formula 
\begin{equation*} se[k] = \sqrt{1 + \frac{1}{B}} \times sd[k]. \end{equation*}

\end{itemize}

We then choose the optimal $k$ to be the smallest value of $k$ such that 
\begin{equation*} Gap[k] \geq Gap[k + 1] - se[k + 1]. \end{equation*}

\begin{figure}[ht!]
     \begin{center}
        \subfigure[]{
            \includegraphics[width=0.47\textwidth]{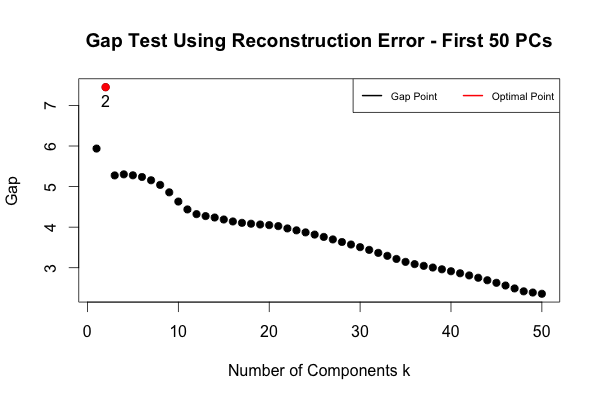}
        }
        \subfigure[]{
           \includegraphics[width=0.47\textwidth]{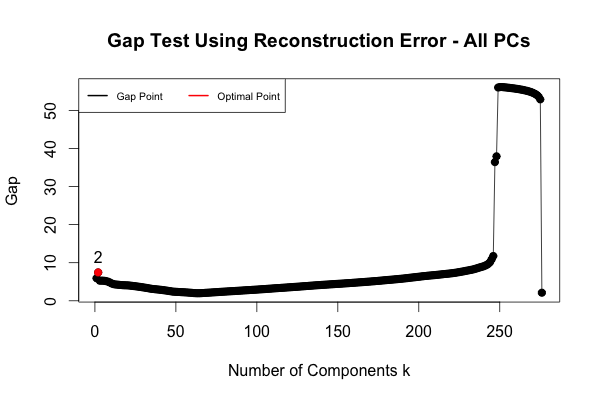}
	}
    \end{center}
    \caption{Gap Statistic (Using Reconstruction Error) on The Number of Principal Components.}
    \label{Gap_RE}
\end{figure}

Figure \ref{Gap_RE} shows a plot of the Gap statistic versus the number of principal components for (a) the first fifty principal components, and (b) the total number of principal components. The optimal number of principal components that minimizes the Gap is $k^* = 2$. 

\subsubsection{Gap Test Using Explained Variation}
In this approach, we define $W_k$ to be the variation explained by the first $k$ principal components. Specifically, $W_k$ is the ratio of the sum of the first $k$ principal components' variances and the sum of the total variance of all principal components. Mathematically speaking,  
\begin{equation*} W_k = \cfrac{\sum_{i = 1}^k V_i}{\sum_{j = 1}^{276} V_j}, \end{equation*}

where $V_i$ is the $i^{th}$ principal component's variance. 

The steps in this Gap test are the same as those for the Reconstruction Error approach with the exception of  $W_k$, which we have described above. We then choose the optimal $k$ to be the smallest value of $k$ such that 
\begin{equation*} Gap[k] \leq Gap[k + 1] - se[k + 1]. \end{equation*}

We note the difference in direction of the two conditions in defining the optimal $k$ between the two Gap-style tests. This is as expected because for the Gap test using reconstruction error, we want to minimize $W_k$ while for the Gap test using explained variation we want to maximize the corresponding $W_k$. However different, this Gap-style test gives the same optimal number of components as the other one which is $k^* = 2$. 

\begin{figure}[ht!]
     \begin{center}
        \subfigure[]{
            \includegraphics[width=0.47\textwidth]{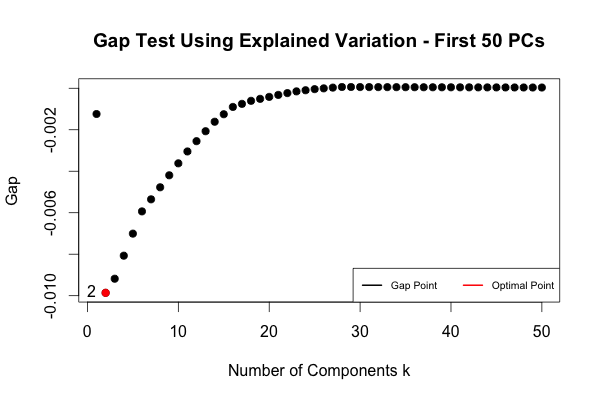}
        }
        \subfigure[]{
           \includegraphics[width=0.47\textwidth]{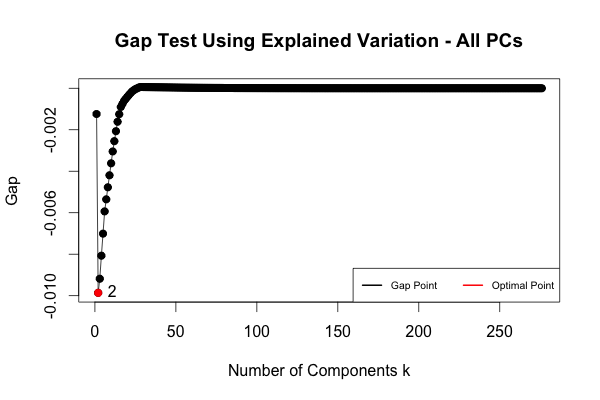}
	}
    \end{center}
    \caption{Gap Statistic (Using Explained Variation) on The Number of Principal Components.}
    \label{Gap_EV}
\end{figure}

Figure \ref{Gap_EV} shows the corresponding plot of the Gap statistic versus the number of principal components for (a) the first fifty principal components, and (b) the total number of principal components.

\subsection{Sparse Autoencoder}

\begin{figure}[ht]
\centering
\includegraphics[scale=.45]{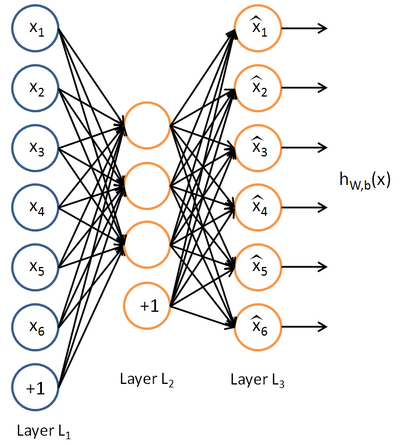}
\caption{Diagram of A $3$-layer Autoencoder Architecture.}
\label{autoencoder_diagram}
\end{figure}

The Autoencoder is based on the concept of sparse coding proposed in a seminal paper by Olshausen et al. $[4]$. In this paper, we implement a $3$-layer Autoencoder (Figure \ref{autoencoder_diagram}) in order to learn a compressed representation (encoding) of the features. Each `neuron' (circle) represents a computational unit that takes as input $x_1$, $x_2$, ...$x_n$ (and a ``$+1$" intercept term, called a bias unit), and outputs 

\begin{equation*}
    h_{W,b}(x) = f\left(W^{(2) T} f\left(W^{(1) T} x + b^{(1)}\right) + b^{(2)}\right)
\end{equation*}

where $f : \mathbb{R} \rightarrow \mathbb{R}$ is called the transfer (activation) function. We choose $f(\cdot)$ to be the hyperbolic tangent (tanh function). The tanh function was chosen instead of the sigmoid function since its output range, [-1,1], more closely approximates the range of our predictor variable than the sigmoid function (range is [0,1]). The tanh activation function is given below:

\begin{equation*}
    f(z) = \frac{\exp(z) - \exp(-z)}{\exp(z) + \exp(-z)}.
\end{equation*}

The leftmost layer of the network is called the input layer, and the rightmost layer the output layer. The middle layer of nodes is called the hidden layer since its values are not observed in the training set. 

The Autoencoder tries to learn a function $h_{W,b}(x) \approx x$. In other words, it is trying to learn an approximation to the identity function, so as to output $\widehat{x}$ that is similar to $x$. The sparse Autoencoder is the Autoencoder with the sparsity constraint added to the objective function. In other words, the objective function of the sparse Autoencoder is given by the reconstruction error with regularization:

\begin{equation*}
    J(W,b) = \frac{1}{m} \sum_{i=1}^{m} \left( \parallel h_{W,b}(x^{(i)}) - x^{(i)} \parallel ^2 \right) + \lambda \sum_{l=1}^{n_l-1} \sum_{i=1}^{s_l} \sum_{j=1}^{s_{l+1}} \left( W_{ji}^{(l)} \right) ^2 + \rho \sum_{l=1}^{n_l-1} \sum_{i=1}^{s_l} \sum_{j=1}^{s_{l+1}} \left|W_{ji}^{(l)}\right|,
\end{equation*}

where $m$ is the number of training samples, $n_l$ is the number of layers ($3$ in our case), and $s_l$ is the number of units in layers $l$. The first term, $J(W, b)$ is an average sum-of-squares error term. The second term is a regularization ($L_2$) term that tends to decrease the magnitude of the weights, and helps prevent overfitting. The weight decay parameter $\lambda$ controls the relative importance of the terms. The sparsity parameter $\rho$ controls how sparse the Autoencoder is. This neural network is then trained using a back-propagation algorithm, where the objective function is minimized using batch gradient descent.

Although the identity function seems like a trivial function to be trying to learn, by placing constraints on the network, such as the weight decay, the sparsity parameter, and the number of hidden units, it is possible to discover interesting structure about the data. When we use a few hidden units, the network is forced to learn a compressed representation of the input. If there is some structure in the data, for example, if some of the input features are correlated, the algorithm will be able to discover some of those correlations, so that the (sparse) Autoencoder often ends up learning a low-dimensional representation similar to PCA.

In order to determine the optimal values of the hyper-parameters and the number of hidden units, we split the data into training, validation and test sets and perform a grid search over the parameter space of the number of units ($[2, 5, 10, 15]$) and the weight decay $\lambda$ ($[0.0001, 0.01, 0.1]$) while using the default value for $\rho$ of $0.01$. The reason for not cross-validating over $\rho$ and for using small sets of possible values for $\lambda$ and the number of hidden units is the high computational cost. For each value of $\lambda$, we use $5$-fold cross-validation on the training data to select the optimal number of hidden units. We then calculate the root mean squared error (RMSE) on the validation data and choose the value for the weight decay corresponding to the smallest RMSE. This way, we obtain the optimal weight decay and the optimal number of units in the hidden layer.

\begin{figure}[ht]
\centering
\includegraphics[scale=.45]{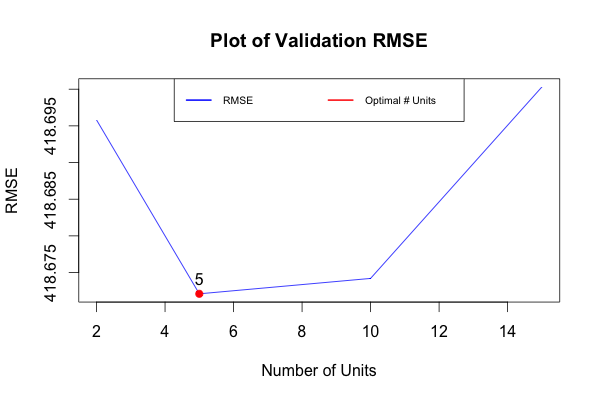}
\caption{Plot of Validation RMSE on The Number of Hidden Units.}
\label{RMSE_Val}
\end{figure}

The optimal weight decay was found to be $\lambda = 0.0001$. For this $\lambda$, we plot the RMSE over the number of units in the hidden layer in Figure \ref{RMSE_Val} and obtain $5$ as the optimal number of units. 

With the determined optimal values of the weight decay, the sparsity parameter, and the number of hidden units, we can also determine the estimated weight matrices and biases. We then fit the training and test data through these weights and biases to obtain the processed data for subsequent analysis. 

\section{Results and Discussion}

We use two models to predict the number of feedbacks: linear regression (a linear model) and regression tree (a non-linear model). We compare the Test RMSE achieved using the output from PCA and sparse Autoencoder for each model. As a baseline, we use the centered and scaled data as input to the models. For PCA, we use the projected data onto the $k^* = 2$ components, and for the sparse Autoencoder, we use the optimal number of $5$ units in the hidden layer, as found by cross-validation. Results are summarized in Table $1$. 

\begin{center}
Table $1$: RMSE on Test Set of Different Models and Methods.  
\end{center}
\begin{center}
\begin{tabular}{| l || c | c |}
\hline
Pre-processing Method & Linear Regression & Regression Tree \\ \hline 
Centering $\&$ Scaling  & $0.8631$ &  $0.6005$ \\ \hline
Principal Component Analysis  & $0.7665$ &  $\mathbf{0.4979}$ \\ \hline
Sparse Autoencoder & $\mathbf{0.5009}$ & $1.0531$ \\ \hline
\end{tabular}
\end{center}

For linear regression, PCA achieves an $11 \%$ improvement compared to the baseline model while the sparse Autoencoder achieves a $42 \%$ improvement. For the regression tree model, PCA achieves a $15 \%$ improvement compared to the baseline model. The sparse Autoencoder, however, performs worse than the baseline model (an increase in RMSE from $0.6005$ to $1.0531$). Unsupervised feature learning in the pre-processing step hence generally improves the prediction accuracy. Taking into account the small range of the outcome variable after scaling and centering, the improvements are certainly significant. 

For the linear regression model the sparse Autoencoder outperforms PCA. This is likely because the sparse Autoencoder solves many of the drawbacks of PCA: PCA only allows linear combinations of the features, restricting the output to orthogonal vectors in feature space that minimize the reconstruction error; PCA also assumes points are multivariate Gaussian, which is most likely not true in many applications, including ours. The sparse Autoencoder is able to learn much more complex, non-linear representations of the data and thus achieves much better accuracy.

An interesting pattern can be observed in the interactions between the linearity and non-linearity of the models and the feature learning methods. The non-linear feature selection method (sparse Autoencoder) achieves significant improvement in RMSE for the linear model (linear regression), while the linear feature selection method (PCA) performs best for the non-linear model (regression tree). Combining the non-linear regression tree model with the non-linear sparse Autoencoder, however, leads to worse results than the baseline. This is likely because regression trees have a top-down construction, splitting at every step on the variable that best divides the data. The sparse Autoencoder returns non-linear combinations of features, making it difficult for regression trees to anticipate. In addition, each tree samples a subset of variables to split on while sparse Autoencoders are known to be sensitive to parameter selection, and hence are not optimized to predict on subsets of the predictor variables, leading to unstable results and poorer performance.

\section{Conclusions and Future Work}

We show empirically that using unsupervised feature learning to pre-process the data can improve the feedback prediction accuracy significantly. These results should be of interest to businesses and publishers in pre-screening or editing their social-media posts prior to publicizing so as to estimate the level of engagement they would be expected to achieve. For instance, an automatic editor may flag posts with low predicted user engagement and draw attention to the writer that revisions may be needed.

Two directions of future work are immediately obvious. First, we can extend the work by trying other unsupervised feature learning methods such as ICA and Kernel PCA in order to better understand how these methods (linear versus non linear) interact with the model type and to what extent our observations can be generalized. Second, we would be interested in investigating whether results can be further improved by using different transfer functions and additional hidden layers in the sparse Autoencoder (i.e., stacked sparse Autoencoder) in order to better capture the time-series aspects of the data set. We may also want to compare the feature learning methods on other models such as SVM, boosting, random forests, etc. \\

\noindent \textbf{Acknowledgements}: We thank Robert Tibshirani for helpful comments.

\subsubsection*{References}
[1] Buza, Krisztian. ``\emph{Feedback prediction for blogs.}" Data Analysis, Machine Learning and Knowledge Discovery. Springer International Publishing, 2014. 145-152.

[2] Cao, K. Chua, W. Chong, H. Lee, and Q. Gu. ``\emph{A comparison of PCA, KPCA and ICA for dimensionality reduction in support vector machine.}" Neurocomputing 55.1 (2003): 321-336.

[3] Robert Tibshirani, Guenther Walther and Trevor Hastie. 
``\emph{Estimating the number of clusters in a data set via the Gap statistic.}" Journal of the Royal Statistical Society, B, 63:411-423, 2001.

[4] Olshausen, Bruno A. ``\emph{Emergence of simple-cell receptive field properties by learning a sparse code for natural images.}" Nature 381.6583 (1996): 607-609.

\end{document}